\documentclass{article}
\usepackage{spconf,amsmath,graphicx,hyperref,listings,booktabs,enumitem,tikz}
\usepackage{algorithm,float}
\usepackage{algpseudocode} 
\usepackage{caption}
\usepackage{pgfplots}
\usepackage{multirow}
\pgfplotsset{compat=1.18}

\usepackage[hypcap=false]{caption}  

\title{From "What to Eat?" to Perfect Recipe: ChefMind's Chain-of-Exploration for Ambiguous User Intent in Recipe Recommendation}
\name{Yu Fu, Linyue Cai, Ruoyu Wu, Yong Zhao*\thanks{This work is funded by the National Natural Science Foundation of China (NSFC) under Grant [No.62177007].}}
\address{Sichuan University\\
        Department of Computer Science\\
        Shuangliu District, Chengdu City, Sichuan Province, China}

\begin{document}
%
\maketitle
\begin{abstract}
Personalized recipe recommendation faces challenges in handling fuzzy user intent, ensuring semantic accuracy, and providing sufficient detail coverage. We propose ChefMind, a hybrid architecture combining Chain of Exploration (CoE), Knowledge Graph (KG), Retrieval-Augmented Generation (RAG), and a Large Language Model (LLM). CoE refines ambiguous queries into structured conditions, KG offers semantic reasoning and interpretability, RAG supplements contextual culinary details, and LLM integrates outputs into coherent recommendations. We evaluate ChefMind on the Xiachufang dataset and manually annotated queries, comparing it with LLM-only, KG-only, and RAG-only baselines. Results show that ChefMind achieves superior performance in accuracy, relevance, completeness, and clarity, with an average score of 8.7 versus 6.4–6.7 for ablation models. Moreover, it reduces unprocessed queries to 1.6\%, demonstrating robustness in handling fuzzy demands. These findings validate ChefMind’s effectiveness and feasibility for real-world deployment.
\end{abstract}
\section{Introduction}
\label{sec:intro}

With the rapid growth of online recipe content, personalized recipe recommendation systems are gaining increasing attention\cite{freyne2010intelligent}. In recent years, various advanced technologie, such as large language models (LLMs), knowledge graphs (KGs), and retrieval-augmented generation (RAG), have been integrated into traditional recommendation frameworks, enhancing system performance. However, these technologies each face significant limitations: LLMs may generate "hallucinations"\cite{ling2023domain}, KGs have limited adaptability in dynamic scenarios\cite{min2022applications}, and RAG heavily relies on retrieval quality and coverage\cite{lewis2020retrieval}. More importantly, existing research primarily focuses on improving individual technologies. How to organically integrate these methods for complementary enhancement remains a critical issue demanding in-depth exploration.

To address these challenges, we propose a novel hybrid architecture that integrates Chain of Enhancement (CoE), Knowledge Graph (KG), and Retrieval-Augmented Generation (RAG) within a unified recipe recommendation system -- ChefMind. The key innovations of our approach are threefold: (1) 
CoE dynamically interprets and refines ambiguous user queries into structured conditions; (2) 
KG provides semantic relational context for accurate and explainable recommendations; and (3) RAG retrieves real-world culinary details to enrich response practicality. This paper systematically quantifies the contribution of each module through extensive experiments and validates the overall superiority of our architecture compared to several baseline methods. The findings demonstrate the efficacy of combining CoE, KG, and RAG to achieve a more robust, intuitive, and personalized recipe recommendation system.

\section{Related Work}
\label{sec:format}

The development of recipe recommendation systems has transitioned from basic matching to semantic-aware understanding. Early systems primarily employed content-based and collaborative filtering\cite{freyne2010recommending,min2019food}. Content-based methods used ingredient and nutritional profiles to suggest similar recipes, offering transparency but struggling to infer implicit preferences\cite{lin2014content,bianchini2017prefer,padmavathi2023recipemate}. Collaborative filtering analyzed user interactions-such as ratings and saves-via models like matrix factorization or neural networks, effectively capturing collective patterns but remaining vulnerable to the cold-start problem\cite{berkovsky2010group,khan2019personalized,rostami2023novel}. Hybrid models were subsequently proposed to integrate both content and interaction signals, improving overall recommendation performance\cite{freyne2010intelligent,pecune2020recommender}. Knowledge graphs (KG) were also introduced to model culinary relationships between ingredients, cuisines, and nutritional attributes\cite{haussmann2019foodkg,chen2021personalized,li2023health}. Using graph neural networks, these systems supported cross-entity reasoning and provided explainable recommendations, thereby enhancing semantic understanding of recipes and dietary contexts\cite{gao2022food}. In recent years, large language models (LLM) and retrieval-augmented generation (RAG) frameworks have been applied, where a retrieval component fetches relevant recipes from large corpora, and the generative language model produces personalized and context-aware suggestions\cite{mohbat2025kerl,wang2024searching}. This approach improves performance on ambiguous queries and balances creativity with accuracy. While some studies suggest combining knowledge graphs with RAG\cite{cai2025practices,cai2024bringing}, a deeper integration of these architectures-particularly within recipe recommendation-remains an open research area.

\section{Methology}
\label{sec:pagestyle}

\subsection{System Architecture}
\label{sys}
\begin{figure}[ht] 
    \centering
    \includegraphics[width=0.5\textwidth]{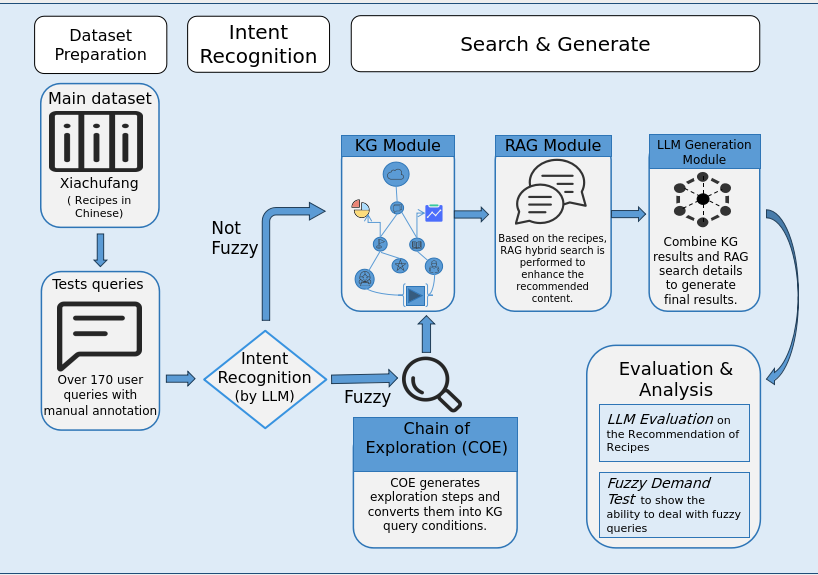} 
    \caption{The Framework of ChefMind}
    \label{fig:recipe_double}
\end{figure}

To address the limitations of traditional recipe recommendation methods, this study proposes ChefMind, a hybrid recommendation system integrating CoE, KG,RAG, LLM. The core objectives are verifying effectiveness advantages and efficiency feasibility, and quantifying individual module contributions using the Xiachufang dataset.

This architecture consists of four complementary modules forming an end-to-end recommendation loop. The framework of our model is shown by figure \ref{fig:recipe_double}. CoE acts as the entry point for fuzzy demand parsing, converting abstract user inputs into quantifiable screening conditions. KG enables semantically accurate recipe screening using structured data stored in Neo4j. RAG supplements unstructured details by encoding corpus into 768-dimensional vectors stored in Milvus for similarity retrieval. LLM functions as an "integrator," combining structured KG results and unstructured RAG details into user-friendly recommendations.

The workflow operates conditionally: for fuzzy demands, CoE generates refinement logic converted into KG query conditions; for clear demands, KG directly processes queries. RAG retrieves the most relevant fragments for candidate recipes, and LLM integrates recipe names with RAG details to generate final natural language recommendations.
Two key formulas are introduced to formalize critical processes of the architecture:



\subsection{CoE Module}
The CoE module serves as the system's intelligent frontend, handling user query parsing through five-level progressive search logic: exact name matching for specific dish queries, ingredient similarity matching based on available ingredients, quick home-style dish retrieval, cuisine and flavor matching for culinary preferences, and broad keyword matching for comprehensive coverage.

This layered strategy precisely responds to specific queries while flexibly handling ambiguous requests. Search results are integrated via matching scores with deduplication ensuring diversity. The CoE module demonstrates its core role in understanding user intent and translating queries into structured conditions for knowledge graph retrieval, forming a complete recommendation loop.

CoE identifies fuzzy demands based on two criteria: the input contains ambiguous terms or has a length of fewer than 5 characters. This judgment logic is mathematically expressed as:
\begin{equation}
\text{fuzzy}(Q) = 
\begin{cases} 
1, & \text{if ambiguous terms} \in Q \lor |Q| < 5 \\
0, & \text{otherwise}
\end{cases}
\label{eq:fuzzy_judgment}
\end{equation}
where $Q$ denotes the user's input query. $\text{fuzzy}(Q)=1$ indicates a fuzzy demand (triggering CoE's 3-step refinement logic), while $\text{fuzzy}(Q)=0$ indicates a clear demand (directly entering KG screening).
\subsection{KG Module}

\begin{figure}
    \centering
    \includegraphics[width=1\linewidth]{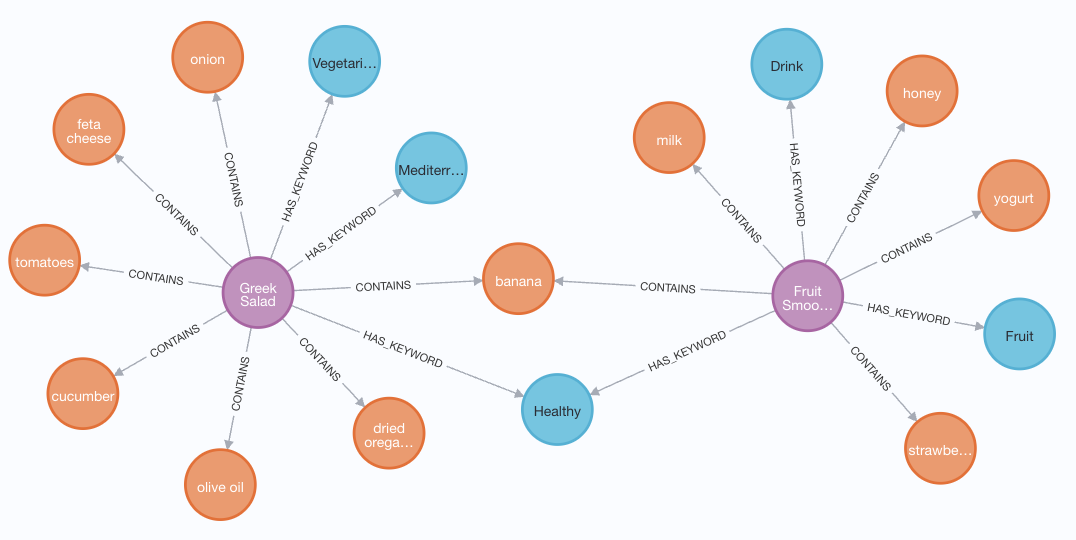}
    \caption{Knowledge graph structure example showing two interconnected recipes with their ingredients and keywords}
    \label{fig:kg_structure}
\end{figure}

The KG module provides semantic and structured recipe data storage and retrieval support, serving as the semantic hub of the recommendation system. Built on the Neo4j graph database, this module comprises three core node types: Recipe, Ingredient, and Keyword, forming a rich semantic network through relationships such as "CONTAINS" and "HAS\_KEYWORD." Recipe nodes include attributes like name, dish type, preparation steps, and author; Ingredient nodes record standardized names, quantities, and units; Keyword nodes manage tags such as "home-style" and regional cuisines. Figure \ref{fig:kg_structure} illustrates a concrete example of this knowledge graph structure, showing two interconnected recipes with their respective ingredients and keywords.

Upon receiving structured query conditions from the CoE module, the KG module employs multi-hop graph traversal and semantic matching to retrieve candidate results satisfying users' multi-dimensional constraints. Leveraging graph database advantages, the system supports real-time queries under complex semantic constraints, achieving high-accuracy retrieval with low latency. The KG module provides complete retrieval paths and associated attributes for each result, enhancing recommendation interpretability and transparency.

\subsection{RAG Module}



The RAG module provides dense vector-based recipe retrieval, serving as the semantic retrieval engine. Built on the Milvus vector database, it transforms recipe content into high-dimensional semantic representations that capture relationships beyond traditional keyword matching.

Each recipe is encoded into dense vectors using pre-trained embedding models. The vector database employs Inner Product metric for similarity computation, achieving sub-millisecond query latency through approximate nearest neighbor search.

The module excels at handling fuzzy demands such as "healthy comfort food" that keyword-based approaches cannot process effectively. It integrates with the KG module through hybrid search strategy, where vector-based retrieval provides broad semantic matching while graph-based retrieval ensures structured constraints, significantly enhancing recommendation accuracy.

RAG retrieves relevant text fragments by calculating vector similarity using cosine similarity:
\begin{equation}
\text{Sim}(\mathbf{v}_q, \mathbf{v}_d) = \frac{\mathbf{v}_q \cdot \mathbf{v}_d}{|\mathbf{v}_q| \cdot |\mathbf{v}_d|}
\label{eq:rag_similarity}
\end{equation}
where $\mathbf{v}_q$ is the 768-dimensional vector of a candidate recipe, $\mathbf{v}_d$ is the vector of a text fragment in the RAG corpus, and higher similarity values indicate greater relevance.

\subsection{LLM Module}
The LLM module functions as the core connection layer of ChefMind, integrating structured data from KG and unstructured details from RAG into coherent recommendations. The module adopts the DeepSeek model for its balanced performance and resource efficiency under GPU constraints.

As the connecting component, LLM performs three key functions: 
(1) integrating KG's candidate recipe names and RAG's contextual details to avoid fragmented information; 
(2) generating natural language output containing recipes and relevant reasons; 
(3) adapting expression to demand types by emphasizing condition matching for fuzzy demands and direct response for clear demands.

The integration process is formalized as:
\begin{equation}
R_{\text{final}} = \text{DeepSeek}(R_{\text{KG}}, D_{\text{RAG}}, P)
\label{eq:llm_integration}
\end{equation}

where $R_{\text{KG}}$ is KG's candidate recipes, $D_{\text{RAG}}$ is RAG's details, $P$ is the demand-based prompt, and $R_{\text{final}}$ is the integrated recommendation output.

LLM bridges data processing modules (CoE, KG, RAG) and end users, converting discrete data into natural language recommendations.

\section{Experiments And Results}
\label{sec:typestyle}
\subsection{Experimental Setup}
The core dataset for this experiment is the "Xiachufang" Chinese recipe dataset. It comprises hundreds of thousands of authentic user-submitted recipes, including structured and unstructured information such as dish names, ingredient lists, detailed cooking instructions, user ratings, and reviews. This dataset closely aligns with the dietary habits and linguistic expressions of Chinese users.

To support model evaluation, we constructed a test set of human-annotated queries, including explicit requests and fuzzy requests, enabling a comprehensive assessment of model performance across different scenarios.

\subsection{Detailed Introduction to Ablation Models}
\textbf{LLM+KG Model}
The core components of this model are LLM and KG Its workflow is as follows: it directly receives user queries, retrieves candidate recipes that meet the conditions through multi-hop graph traversal and semantic matching of Knowledge Graph based on the relationship between Recipe, Ingredient, and Keyword nodes, and then the Large Language Model organizes the structured recipe information returned by Knowledge Graph, including dish names, ingredients, and steps, into natural language recommendations.

\textbf{LLM+RAG Model}
The core components of this model are LLM and RAG. Its workflow is as follows: it encodes user queries and recipe texts intovectors, retrieves the most relevant unstructured text fragments including step details and cooking tips through cosine similarity calculation in Milvus, and the Large Language Model generates recommendations based on the retrieved results.

\textbf{ChefMind}
The core components of this model are LLM, CoE, KG and RAG. Its funtion has been mentioned in \ref{sys}

\subsection{Scoring Criteria}
\textbf{}{Scoring Criteria Design}
The experiment adopts a four-dimensional quantitative scoring system, with LLM serving as an objective evaluator to score the recommendation results generated by the models on a scale of 1 to 10, where 10 is the highest score. The total score is the average of the four dimensions, retaining one decimal place. The specific criteria are as follows:

To evaluate the recommendation results, we designed a four-dimensional scoring system, where each dimension is rated on a scale of 1 to 10 and the final score is the average value:
\begin{itemize}[noitemsep, topsep=0pt]
    \item \textbf{Accuracy} – measures the degree of consistency between the recommendation results and the ground-truth labels. Higher scores indicate better alignment with user expectations.
    \item \textbf{Relevance} – assesses how closely the recommendations match the intent of the user’s query while avoiding irrelevant or off-topic suggestions.
    \item \textbf{Completeness} – evaluates the coverage of essential information, including dish names, ingredients, preparation steps, and contextual scenarios. Missing items result in deductions.
    \item \textbf{Clarity} – reflects the logical structure, readability, and linguistic standardization of the recommendations. Poorly ordered steps, inconsistent terms, or chaotic expressions reduce the score.
\end{itemize}

\subsection{Experimental Result Analysis}

The experimental result is that the performance of ChefMind is significantly better than that of the ablation models (shown in Table \ref{tab:ablation_experiment_batch_results}. In terms of the overall average total score, ChefMind (8.7 points) is higher than LLM+RAG (6.7 points) and LLM+KG (6.4 points). ChefMind leads by 2 to 3 points in the accuracy and relevance dimensions, which reflects the improvement of Chain of Exploration on intent parsing and the collaborative value of Knowledge Graph and Retrieval-Augmented Generation. In terms of the number of unprocessed queries, ChefMind has only 2 unprocessed queries, accounting for 1.6\%, which is much lower than LLM+KG (33 unprocessed queries, accounting for 25.6\%) and LLM+RAG (22 unprocessed queries, accounting for 17.1\%). Especially in batches with fuzzy queries, such as batches 3 and 7, ChefMind has only 1 unprocessed query, while LLM+KG has 4 to 5 unprocessed queries.

\begin{center}
\centering
\captionof{table}{Batch-Level Result Comparison of ChefMind Ablation Experiment (13 Batches, 129 Queries in Total)}
\resizebox{\linewidth}{!}{
\begin{tabular}{c c c c c c c c c}
\toprule
\multirow{2}{*}{\textbf{Batch}} & \multirow{2}{*}{\textbf{Total Queries}} & 
\multicolumn{2}{c}{\textbf{LLM+KG}} & 
\multicolumn{2}{c}{\textbf{LLM+RAG}} & 
\multicolumn{2}{c}{\textbf{ChefMind}} \\
\cmidrule(lr){3-4} \cmidrule(lr){5-6} \cmidrule(lr){7-8}
& & \textbf{Avg Score} & \textbf{Unprocessed Queries} & \textbf{Avg Score} & \textbf{Unprocessed Queries} & \textbf{Avg Score} & \textbf{Unprocessed Queries} \\
\midrule
1 & 10 & 6.2 & 3 & 6.5 & 2 & 8.8 & 0 \\
2 & 10 & 6.5 & 2 & 6.8 & 1 & 8.9 & 0 \\
3 & 10 & 5.8 & 5 & 6.0 & 4 & 8.2 & 1 \\
4 & 10 & 6.7 & 2 & 7.0 & 1 & 9.0 & 0 \\
5 & 10 & 6.3 & 3 & 6.6 & 2 & 8.7 & 0 \\
6 & 10 & 6.9 & 1 & 7.2 & 1 & 9.1 & 0 \\
7 & 10 & 5.9 & 4 & 6.2 & 3 & 8.3 & 1 \\
8 & 10 & 6.6 & 2 & 6.9 & 1 & 8.9 & 0 \\
9 & 10 & 6.4 & 3 & 6.7 & 2 & 8.6 & 0 \\
10 & 10 & 6.8 & 1 & 7.1 & 1 & 9.0 & 0 \\
11 & 10 & 6.5 & 2 & 6.8 & 1 & 8.8 & 0 \\
12 & 10 & 6.1 & 3 & 6.4 & 2 & 8.5 & 0 \\
13 & 9 & 6.3 & 2 & 6.6 & 1 & 8.7 & 0 \\
\midrule
\textbf{Overall} & \textbf{129} & \textbf{6.4} & \textbf{33} & \textbf{6.7} & \textbf{22} & \textbf{8.7} & \textbf{2} \\
\bottomrule
\end{tabular}
}
\vspace{1mm}
\footnotesize{Note: 1. Avg Score is the average of the four-dimensional scores, retaining one decimal place; 2. Unprocessed Queries is the number of unprocessed queries in the batch, unit: piece; 3. Overall is the summary result of 13 batches.}
\label{tab:ablation_experiment_batch_results}
\end{center}



\begin{figure}[htbp]
\centering
\begin{tikzpicture}
\begin{axis}[
    ybar,
    bar width=8pt, 
    width=0.48\textwidth, 
    height=6cm,
    ymin=0, ymax=10,
    ylabel={\small Score (1-10)},
    xlabel={\small Dimension},
    xtick=data,
    xticklabels={Acc, Rel, Comp, Clarity, Overall},
    x tick label style={font=\small, rotate=45, anchor=east},
    legend style={
        font=\footnotesize,
        at={(0.5,-0.35)},
        anchor=north,
        legend columns=3,
        /tikz/every even column/.append style={column sep=0.3cm}
    },
    yticklabel style = {font=\small},
    ylabel style = {font=\small},
    xlabel style = {font=\small},
    nodes near coords={\pgfmathprintnumber[precision=1]{\pgfplotspointmeta}}, 
    nodes near coords align={vertical},
    every node near coord/.append style={font=\tiny, yshift=2pt},
    enlarge x limits=0.15,
    symbolic x coords={Acc, Rel, Comp, Clarity, Overall},
    axis lines*=left, 
    ymajorgrids=false, 
]

\definecolor{dopamine1}{HTML}{FF6B9D} 
\definecolor{dopamine2}{HTML}{72DDF7} 
\definecolor{dopamine3}{HTML}{97FEED} 

\addplot[fill=dopamine1!80, draw=black, very thin] coordinates {
    (Acc, 6.2) (Rel, 6.0) (Comp, 6.5) (Clarity, 6.8) (Overall, 6.4)
};
\addplot[fill=dopamine2!80, draw=black, very thin] coordinates {
    (Acc, 6.5) (Rel, 6.3) (Comp, 7.0) (Clarity, 7.1) (Overall, 6.7)
};
\addplot[fill=dopamine3!80, draw=black, very thin] coordinates {
    (Acc, 8.5) (Rel, 8.8) (Comp, 8.6) (Clarity, 9.0) (Overall, 8.7)
};

\legend{LLM+KG, LLM+RAG, ChefMind}
\end{axis}
\end{tikzpicture}
\caption{Performance comparison across dimensions. ChefMind (mint green) consistently outperforms baselines.}
\label{fig:perf_compact}
\end{figure}
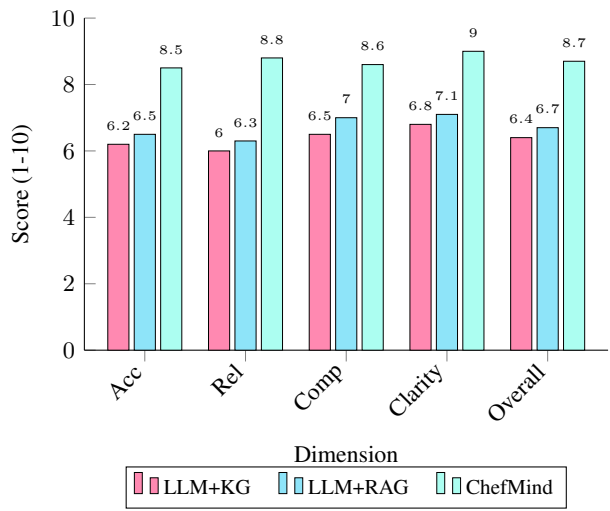

As shown in Table~\ref{tab:ablation_experiment_batch_results}, ChefMind consistently outperforms the ablation models. This superiority is further visualized in Figure~\ref{fig:perf_compact}, which breaks down the average scores by evaluation dimension.

\section{Conclusion}
\label{sec:majhead}

This paper proposes a hybrid CoE+KG+RAG+LLM architecture, ChefMind, for recipe recommendation in Chinese. Experimental results demonstrate that the architecture outperforms LLM+KG and LLM+RAG models in accuracy, relevance, completeness and clarity, especially in fuzzy demands.

\bibliographystyle{IEEEbib}
\bibliography{strings,refs}

\end{document}